# Lung Nodule Classification using Deep Local-Global Networks

**Mundher Al-Shabi[a, 1], Boon Leong Lan[a], Wai Yee Chan[b], Kwan-Hoong Ng[b], Maxine Tan[a, c]**

a) Electrical and Computer Systems Engineering Discipline, School of Engineering, Monash University Malaysia, 47500 Bandar Sunway, Selangor, Malaysia
b) Department of Biomedical Imaging, University of Malaya, 50603 Kuala Lumpur, Malaysia
c) School of Electrical and Computer Engineering, University of Oklahoma, Norman, OK 73019, USA

## Abstract

**Purpose**: Lung nodules have very diverse shapes and sizes, which makes classifying them as benign/malignant a challenging problem. In this paper, we propose a novel method to predict the malignancy of nodules that have the capability to analyze the shape and size of a nodule using a global feature extractor, as well as the density and structure of the nodule using a local feature extractor.

**Methods**: We propose to use Residual Blocks with a 3×3 kernel size for local feature extraction, and Non-Local Blocks to extract the global features. The Non-Local Block has the ability to extract global features without using a huge number of parameters. The key idea behind the Non-Local Block is to apply matrix multiplications between features on the same feature maps.

**Results**: We trained and validated the proposed method on the LIDC-IDRI dataset which contains 1,018 computed tomography (CT) scans. We followed a rigorous procedure for experimental setup namely, 10-fold cross-validation and ignored the nodules that had been annotated by less than 3 radiologists. The proposed method achieved state-of-the-art results with AUC=95.62%, while significantly outperforming other baseline methods.

**Conclusions**: Our proposed Deep Local-Global network has the capability to accurately extract both local and global features. Our new method outperforms state-of-the-art architecture including Densenet and Resnet with transfer learning.

## 1. Introduction

Deep learning based models were reasonably trained for lung nodule classification [1–4] with area under the receiver operating characteristic curve (AUC) values of 87.7% or less[3]. However, handling local and global features is very challenging. Lung nodules have diverse shapes and sizes [5], which require global feature extraction that can describe the nodule shape and size as a whole. On the other hand, local features are important for lung nodule classification to pay attention to details including the nodule density and texture.

The standard way to capture global and local features in the deep neural network is applying FC (full connections) to all layers. That is, every neuron in any layer should be connected to all previous neurons in the previous layer. Despite its capability to capture global and local features, FC requires an exponential number of weights/parameters. It is thus common to only apply a FC layer in the last stage of the neural network while all other layers are designed in the form of CNNs (convolutional neural networks) [6, 7]. The CNN requires a small number of parameters compared to FC, as a group of parameters (the kernel) in CNN is repeated/shared and applied over all other features. However, CNN fails to detect global features due to its small kernels. One way to overcome this limitation is to increase the kernel size, but this requires more parameters.

---

[1] Corresponding author, email: mundher.al-shabi @ monash.edu

An alternative solution is to use dilated convolution [8], whereby the kernel size can be increased without increasing the number of parameters. Although the dilated convolution solves the multi-scale problem by increasing or decreasing the kernel size, we still need to manually specify a pre-defined dilation rate. The author manually defined two dilation rates previously[8]: The first dilation rate of 1 is suitable for smaller nodules and the second dilation rate of 2 is suitable for bigger nodules. That is, to cover bigger input receptive fields, we used another dilated convolution.

In this study, we propose to apply residual convolution [6] to detect small, local features and non-local convolution [9, 10] for multi-sized global features. Overall, our contributions in this work can be summarized as follows:

1. We proposed using a novel method called Local-Global neural network for lung nodule classification
2. We studied other variations of Local-Global neural network as well as transfer learning and state-of-the-art models including Densenet and Resnet
3. Our proposed method outperforms the baseline methods and state-of-the-art models on the public Lung Image Database Consortium image collection (LIDC-IDRI) dataset with an AUC of 95.62%

## 2. Methods

Residual Block, first introduced in Resnet [6] has become a de facto standard block in designing CNNs for classifying lung nodules [4, 11]. A significant feature of the Residual Block is the residual skip connection – see Figure 1. The skip connection allows gradients to pass freely to the lower layers, which prevents the gradient from vanishing or exploding. The Residual Block has a 3×3 kernel size, which is fast to compute and has a lower number of parameters. As the Residual Block has a small receptive input area, it is useful for extracting local features.

To extract the global features, we used Non-Local neural networks [9], also known as Self-Attention layers [10]. As shown in Figure 1, the Non-Local Block starts with a linear convolution of the input features. Each of these convolutions has a kernel size of 1×1. The results of this linear transformations are $f(x)$, $g(x)$ and $h(x)$. Then, we applied the core function/idea of Non-Local networks, which is basically matrix multiplications between the features as shown in Equation 1:

$$x + h(x)\, Softmax(f(x)^T g(x))^T \gamma \qquad (1)$$

The advantage of using matrix multiplication is to create a non-linear interaction between multiple features of overall spatial locations. Each feature in $f(x)$ is multiplied by all features in $g(x)$. This allows the network to capture global features without the need for extra parameters as matrix multiplication is a parameter-free operation. Then we applied the Softmax function to the output of the matrix multiplication. Each feature after Softmax is between 0 and 1 and the sum of each of the feature maps is 1. These feature maps act as attention masks, whereby they allow the network to attend to particular regions in $h(x)$. After multiplying the output of the Softmax function by $h(x)$, the regions with zero attention will be removed and will not be passed to the next layer.

In this study, we follow self-attention implementation [10] of the Non-Local network, whereby we multiplied the result by a scalar value $\gamma$, which a learnable parameter and set to zero by default [10]. The $\gamma$ value regulates/controls the contribution of the Non-Local layer to the overall output. For example, if $\gamma$ is zero, the whole Non-Local feature contribution will be zero, and input $x$ will be passed through instead. After each of the Non-Local Blocks, we applied dropout regularization to improve the generalization of the network.

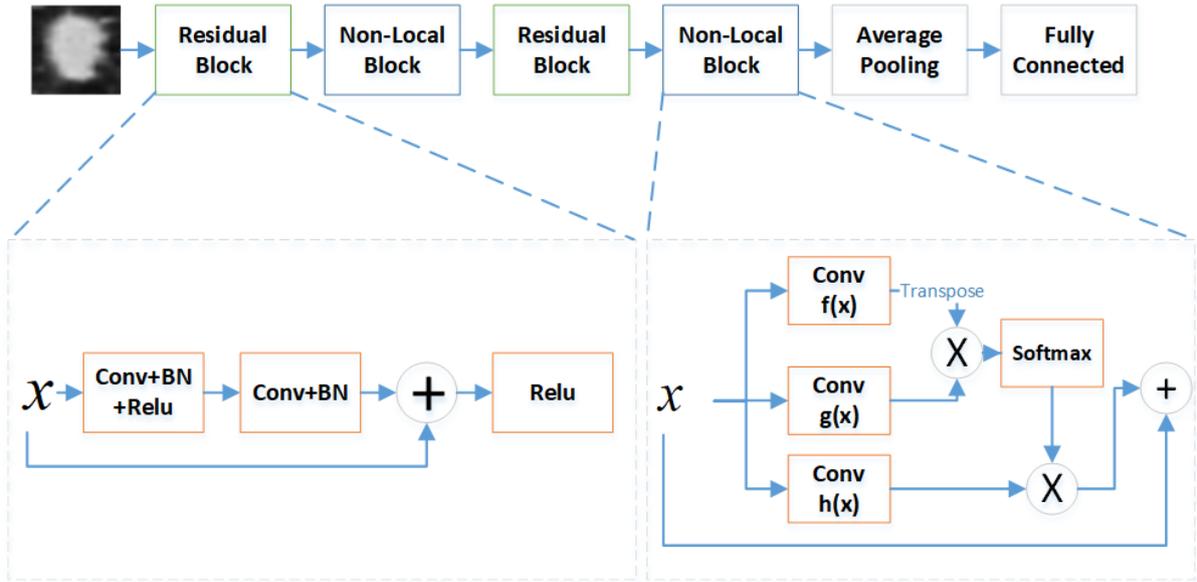

*Figure 1: The overall architecture of the proposed Local-Global method. 'X' denotes matrix multiplication and '+' elementwise addition; BN denotes batch normalization.*

As shown in Figure 1, the Local-Global network has two Residual Blocks and two Non-Local Blocks, and the output of each of these blocks has 32 feature maps or channels. Moreover, we kept the spatial size (width and height) of each feature map at 32×32 by adding padding before the convolution. After that, we applied global average pooling with a kernel size of 32. This average pooling takes the average of each feature map, which results in a vector of size 32. Finally, we passed this vector to a Fully-Connected single neuron with a sigmoid activation function as shown in Figure 1 and Table 1.

*Table 1: Local-Global network and other baseline method architectures. 'CH' denotes channels or number of filters, '–'denotes no operation, and 'P' is the dropout probability.*

| BasicResnet | Local-Global | AllAtn | AllAtnBig |
|---|---|---|---|
| Residual Block, 32 CH | | Conv, 32 CH | |
| -- | Non-Local Block, 32 CH | | |
| Dropout, $P$=0.1 | | | |
| Residual Block, 32 CH | | -- | Non-Local Block, 32 CH |
| -- | Non-Local Block, 32 CH | | |
| Dropout, $P$=0.2 | | | |
| Average Pooling, kernel size=32 | | | |
| Fully Connected, single neuron | | | |

## 2.1. Baseline Methods

We designed multiple baseline methods and compared them with our proposed method. First, we used state-of-the-art architectures for computer vision tasks, namely Densenet121 [12], and Resnet [6] with two variations i.e., Resnet50 and Resnet18. These three networks were pre-trained using ImageNet dataset. Then, we replaced the fully-connected layer of size 1000 with a single neuron and sigmoid activation function. We only trained the last layer and kept all other weights fixed. This procedure was applied to improve the network generalization taking into account that we had a small number of training instances. These models accept inputs with three channels, therefore we copied the same nodule image and applied them to the three channels.

The fourth baseline method is BasicResnet network (in Table 1), which is similar to the Local-Global Network, but without the Non-Local Block. This baseline method allows us to analyze the effectiveness of the Non-Local Block.

The fifth and sixth baseline methods, namely AllAtn and AllAtnBig contain only Non-Local Blocks. These two baseline methods are useful to understand the importance of the local block (Resnet/Residual Block).

## 2.2. Experimental Setup

We trained the networks using the LIDC-IDRI dataset [13]. We followed the same rigorous procedure for data pre-processing explained in [8], except that we classified the nodule directly without using a nodule segmentation mask.

Our proposed method and all the baseline methods were trained for 50 epochs using Adam optimizer [14] with learning rate set to 0.001 and batch size of 256. As this a binary classification problem, we chose Binary Cross Entropy as the loss function. All the experiments were designed and written in Python using the PyTorch framework and were run on an Nvidia RTX 2080 Ti GPU.

We used the standard tool ROCKIT$^{TM}$ [15] to calculate the AUC on the test data in all 10 folds of the 10-fold cross-validation experiment. Similarly, the accuracy is calculated using equation (2) below:

$$Accuracy = \frac{TP + TN}{TP + TN + FP + FN} \quad (2)$$

We also calculated the precision and the sensitivity of our model on the test data using equations (3) and (4) below:

$$Precision = \frac{TP}{TP + FP} \quad (3)$$

$$Sensitivity = \frac{TP}{TP + FN} \quad (4)$$

## 3. Results

As shown in Figure 2 and Table 2, the Local-Global network outperforms other methods on all metrics except the sensitivity, whereby BasicResnet outperforms the Local-Global network by only 0.0026. In terms of AUC, our method significantly outperformed the other baseline methods by a considerable/high margin.

*Table 2: Classification performance comparisons of the proposed Local-Global method against other baseline methods on the LIDC-IDRI dataset.*

| Methodology | AUC | Accuracy | Precision | Sensitivity |
|---|---|---|---|---|
| Local-Global | 95.62% | 88.46% | 0.8738 | 0.8866 |
| BasicResnet | 93.42% | 85.63% | 0.8242 | 0.8892 |
| AllAtn | 86.18% | 78.80% | 0.7616 | 0.8103 |
| AllAtnBig | 85.89% | 77.97% | 0.8137 | 0.6995 |
| Resnet50 | 86.82% | 77.62% | 0.8016 | 0.7069 |
| Resnet18 | 86.41% | 78.21% | 0.7855 | 0.7488 |
| Densenet121 | 92.50% | 84.57% | 0.8686 | 0.7980 |

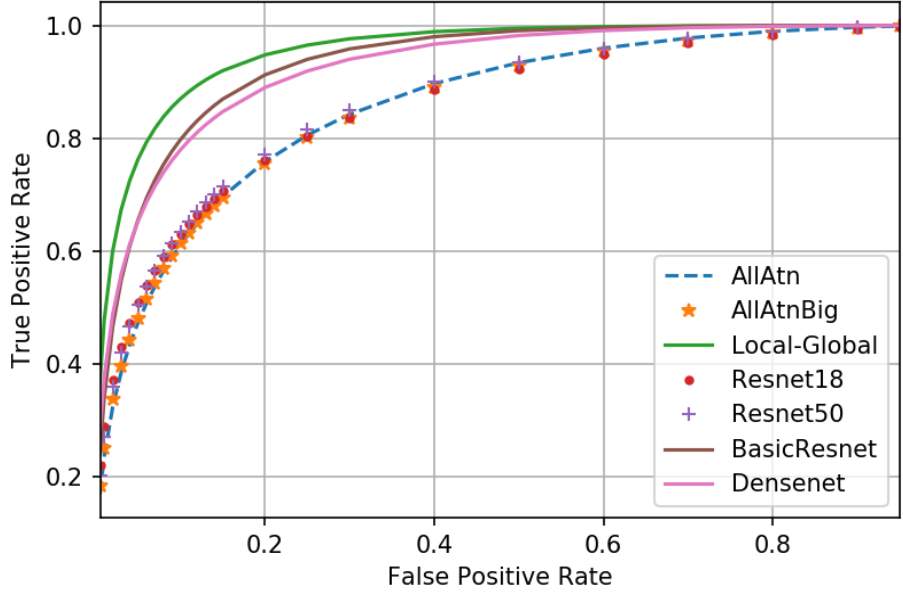

*Figure 2: The receiver operating characteristic (ROC) curve of our Local-Global network compared with 5 other baseline methods. It can be observed that the ROC curve of the Local-Global network significantly outperforms the other methods.*

The state-of-the-art methods including Resnet50 and Resnet18 performed poorly even when we applied transfer learning. Although Densenet achieved 92.50% AUC, it is lower than BasicResnet and Local-Global network in terms of accuracy, sensitivity and AUC. Transfer learning does not work perfectly when pre-training is conducted on non-medical datasets; this limitation is clear when we compare transfer learning with fine-crafted network design, such as Local-Global networks.

Additionally, the local feature extractor based network, namely BasicResnet significantly outperformed the global feature extractors, such as AllAtn and AllAtnBig. This shows that extracting the long dependencies in the Non-Local Block alone is not enough, but it would be useful if we combine it with local features as we have done with the Local-Global network.

## 4. Conclusions

In this study, we proposed Local-Global networks that have the capacity to accurately extract both local and global features well for lung nodule classification. Our new method considerably outperforms six other baseline methods including state-of-the-art architectures, such as Densenet and Resnet with transfer learning. The results show that our Local-Global network extracts multi-scale features and can generalize well across unseen data.

## Acknowledgments

This work was supported by the Fundamental Research Grant Scheme (FRGS), Ministry of Education Malaysia (MOE), under grant FRGS/1/2018/ICT02/MUSM/03/1.

## Compliance with Ethical Standards

Funding: This study was funded by the Fundamental Research Grant Scheme (FRGS), Ministry of Education Malaysia (MOE), under grant FRGS/1/2018/ICT02/MUSM/03/1.

Conflict of Interest: The authors declare that they have no conflict of interest.

Ethical approval: All procedures performed in studies involving human participants were in accordance with the ethical standards of the institutional and/or national research committee and with the 1964 Helsinki declaration and its later amendments or comparable ethical standards.

Informed consent: Informed consent was obtained from all individual participants included in the study.